\pdfoutput=1
\documentclass{article}

\usepackage[final]{nips_2016}

\usepackage[utf8]{inputenc} 
\usepackage[T1]{fontenc}    
\usepackage{url}            
\usepackage{booktabs}       
\usepackage{amsfonts}       
\usepackage{nicefrac}       
\usepackage{microtype}      

\usepackage{amsmath}
\usepackage{amssymb}
\usepackage{amsfonts}
\usepackage{amsthm}
\theoremstyle{definition}
\newtheorem{remark}{Remark}
\usepackage{shortcuts}
\usepackage{graphicx}
\usepackage{caption}
\usepackage{subcaption}
\graphicspath{{./images/}}

\usepackage{algorithm}
\usepackage{algorithmic}

\title{Decentralized Topic Modelling\\with Latent Dirichlet Allocation}

\newcommand*\samethanks[1][\value{footnote}]{\footnotemark[#1]}

\author{
  Igor Colin \thanks{The two authors contributed equally.}\\
  LTCI, Télécom ParisTech\\
  Université Paris-Saclay, 75013 Paris
  \And
  Christophe Dupuy \samethanks\\
  Technicolor; INRIA - Sierra Project - Team\\
  2, rue Simone Iff, 75012 Paris
}

\begin{document}

\maketitle

\begin{abstract}
Privacy preserving networks can be modelled as decentralized networks (\eg sensors, connected objects, smartphones), where communication between nodes of the network is not controlled by a master or central node. For this type of networks, the main issue is to gather/learn global information on the network (\eg by optimizing a global cost function) while keeping the (sensitive) information at each node. 
In this work, we focus on text information that agents do not want to share (\eg, text messages, emails, confidential reports). We use recent advances on decentralized optimization and topic models to infer topics from a graph with limited communication.
We propose a method to adapt latent Dirichlet allocation (LDA) model to decentralized optimization and show on synthetic data that we still recover similar parameters and similar performance at each node than with stochastic methods accessing to the whole information in the graph.


\end{abstract}

\section{Introduction}
\label{sec:introduction}
Decentralized networks, \ie networks with limited communication between nodes, provide an ideal framework for privacy preserving optimization. They are particularly adapted to the case where agents seek to minimize a global cost function which is separable in the data collected locally by each agent. In such networks, it is typically impossible to efficiently centralize data or to globally aggregate intermediate results: agents can only communicate with their immediate neighbors, often in a completely asynchronous fashion. In a privacy setting, agents never share the raw information they have but rather communicate aggregated information or parameters. Recent work on decentralized optimization focuses on convex optimization: such methods are mostly based on gradient descent~\cite{nedic2009distributed, nedic2011asynchronous} or dual averaging~\cite{agarwal2010distributed, colin2016gossip} and theoretical convergence upper bounds have been established for both synchronous and fully asynchronous computations.

In this paper, we tackle the non-convex problem of topic modelling, where agents have sensitive text data at their disposal that they can not or do not want to share (e.g., text messages, emails, confidential reports).
More precisely, we adapt the particular Latent Dirichlet Allocation (LDA) \cite{LDA} model to decentralized networks. We combine recent work of \cite{MeOnlineEM2016} on online inference for latent variable models, which adapts online EM~\cite{OnlineEM} with local Gibbs sampling in the case of intractable latent variable models (such as LDA) and recent advances on decentralized optimization \cite{agarwal2010distributed, colin2016gossip}. The method presented in \cite{MeOnlineEM2016} is particularly adapted to decentralized framework as it consists in iteratively updating sufficient statistics, which can be done locally. After presenting our {\sc DeLeDA} (for Decentralized LDA) algorithm, we give a brief sketch of convergence proof. Then, we apply our new method to synthetic datasets and show that our method recovers the same parameters and has similar performance than the online method~\cite{MeOnlineEM2016} after enough iterations.


\section{Background}
\label{sec:background}
\paragraph{Latent Dirichlet allocation (LDA) \citep{LDA}.}
Let $D$ be the number of documents of a corpus ${\mathcal{C}=\{X_1,\ldots,X_D\}}$, $V$ the number of words in our vocabulary and $K$ the number of latent topics in the corpus. Each topic $\beta^k$ corresponds to a discrete distribution on the $V$ words (that is an element of the simplex in $V$ dimensions). A hidden discrete distribution $\theta_d$ over the $K$ topics  (that is an element of the simplex in $K$ dimensions) is attached to each document $d$.
LDA is a generative model applied to a corpus of text documents which assumes that each word of the $d$-th document $X_d$ is generated as follows:
\begin{itemize}
\item Choose $\theta_d\sim\mathrm{Dirichlet}(\alpha)$,
\item For each word $x_n\in X_i=\left(x_1,\ldots,x_{N_{X_d}}\right)$:
\begin{itemize}
\item Choose a topic $z_n\sim\mathrm{Multinomial}(\theta_d)$,
\item Choose a word $x_n\sim\mathrm{Multinomial}(\beta^{z_n})$.
\end{itemize}
\end{itemize}
In LDA, the inference consists in learning the topic matrix $\beta\in\mathbb{R}^{K\times V}$ --- global parameter --- and topic proportions $\theta_d$ of each document --- local variable --- from a corpus of documents. We choose to adapt Gibbs Online EM for LDA (G-OEM) \cite{MeOnlineEM2016} to the decentralized framework. 
The other inference methods for LDA (\eg \cite{LDA,OnlineLDA,GibbsLDA,SinglePassLDA,mimno2012sparse}) are very close to G-OEM but are less robust in practice (see \cite{MeOnlineEM2016} for complete analysis).

\paragraph{Gibbs Online EM for LDA \cite{MeOnlineEM2016}.} We consider an exponential family model on random variables $(X,h)$ with parameter $\eta\in\mathcal{E}\subseteq\mathbb{R}^d$ and with density \citep{ExpFam1998}:
\begin{equation}
p(X,h|\eta) = a(X,h)\exp\big[\langle\phi(\eta), S(X,h)\rangle -\psi(\eta)\big],
\label{eqn:exp_fam}
\end{equation}
with $h$ hidden and $X$ observed.
G-OEM is an adaptation of the online EM algorithm \cite{OnlineEM} that maximizes the likelihood of observed data $\max_{\eta\in\mathcal{E}} \sum_{i=1}^N\log p(X_i|\eta)$ in the case where the conditional $p(h|X,\eta)$ is intractable to compute, such as for LDA. In particular, the G-OEM algorithm iteratively updates the sufficient statistics $s$ of the model through the formula at step $t+1$:
\begin{equation}
s^{t+1} = (1-\rho_{t+1})s^t + \rho_{t+1}\mathbb{E}_{p(h_{t+1}|X_{t+1},\eta^*(s^{t}))}[S(X_{t+1},h_{t+1})],
\label{eqn:G-OEM}
\end{equation}
where $\eta^*(s)$ is the maximum of likelihood for the sufficient statistics $s$, \ie ${\eta^\ast(s) \in \arg\max\  \langle \phi(\eta),s \rangle - \psi(\eta)}$ (which is the usual M-step update in the EM algorithm~\cite{EM1977}). For LDA, the expectation in \eqref{eqn:G-OEM} is intractable to compute and is approximated with Gibbs sampling --- see \cite{MeOnlineEM2016} for more details.

In this setting for LDA, an observation is a document ${X_i}$. Each observation $X_i$ is associated with the hidden variables $h_i$, with ${h_i\equiv(Z_i=(z_1,\ldots,z_{N_{X_i}}),\theta_i)}$.  The vector $\theta_i$ represents the topic proportions of document $X_i$ and $Z_i$ is the vector of topic assignments of each word of $X_i$. The hidden variable $h_i$ is local, \ie attached to one observation $X_i$. The parameters of the model are global, represented by ${\eta\equiv(\beta,\alpha)}$, where $\beta$ represents the topic matrix and $\alpha$ represents the Dirichlet prior on topic proportions.

\paragraph{Decentralized setting}
Let $n > 0$ and let $\mathcal{G} = ([n], \mathcal{E})$ be an undirected graph of $n$ agents. We consider a setting where each agent aims to optimize a global ojective, without revealing its local observations. In addition, we are interested in the case where communication is possibly asynchronous, \ie there is no global clock ensuring iterations synchrony between nodes. Gossip algorithms~\cite{boyd2006a} are particularly adapted to this setting, as they rely on peer-to-peer interactions. Initially, each agent $i \in [n]$ stores an iterates $s_i^0$; at each iteration, a node awakes and select one of its neighbor: the two nodes then average their iterates. Then, in the synchronous (resp. asynchronous) setting, all the nodes (resp. the two active nodes) update their iterate according to a local rule, say~\eqref{eqn:G-OEM}.

When performing decentralized \emph{estimation} \eg estimating the sample mean, the averaging step allows for a quick uniformization of the network iterates. Indeed, nodes iterates converge to the network average at an exponential rate which depends on some communication property of the graph called the spectral gap: the larger the spectral gap, the faster the convergence. In the case of decentralized \emph{optimization}, since the network is not conserving its mass during the process  --- typically, gradients are added ---, the convergence is not guaranteed in general. However, provided that at each iteration the mass added to the network is bounded, one can formulate upper bounds on the convergence rate, usually polynomial in the number of iterations.


\section{{\sc DeLeDA} algorithm}
\label{sec:delada-algorithm}

In this paper, we aim at combining both G-OEM for LDA and decentralized optimization. Since the G-OEM updates are very similar to stochastic gradient updates, the reasoning of decentralized convex optimization can be extended to G-OEM. In a synchronous setting, our algorithm coined {\sc DeLeDA} consists in the following: each node $i \in [n]$ initializes its sufficient statistics to $s_i^0$. At each iteration, the two awaken nodes average their statistics. Then, every node performs a local update of its statistics using update rule~\eqref{eqn:G-OEM} and its own observations. The method is detailed in Algorithm~\ref{alg:deleda}.

\begin{algorithm}[t]
  \caption{Decentralized LDA in the synchronous setting.}
  \label{alg:deleda}
  \begin{algorithmic}[1]
    \REQUIRE Step size $(\rho_t)_{t \geq 1} > 0$.
    \STATE Each node $i \in [n]$ initializes $s_i$ at $s_i^0$.
    \FOR{$t = 1, \ldots, T$}
    \STATE Draw $(i, j)$ uniformly at random from $\mathcal{E}$
    \STATE Set $s_i, s_j \gets \frac{s_i + s_j}{2}$
    \FOR{$k = 1, \ldots, n$}
    \STATE Update (locally) sufficient statistic $s_k$ with \ref{eqn:G-OEM}
    \ENDFOR
    \ENDFOR
    \STATE \textbf{return} Each node $k$ has $s_k$
  \end{algorithmic}
\end{algorithm}

For $t \geq 0$, let us define the average iterate $\overline{s}^t = n^{-1} \sum_{i = 1}^n s_i^t$.
Using Algorithm~\ref{alg:deleda} description and update rule~\eqref{eqn:G-OEM}, one can see that $\overline{s}^{t + 1}$ is obtained from $\overline{s}^t$ using the G-OEM update rule with all observations of the network. Therefore, $\overline{s}^t$ convergence derives from G-OEM theoretical guarantees~\cite{MeOnlineEM2016}; in order to prove the convergence of each agent, one only has to show that for all $i \in [n]$, $s_i^t$ tends to $\overline{s}^t$ as the number of iterations $t$ grows.

Let us define the iterates matrix $\mathbf{S}^t = (s_1^t, \ldots, s_n^t)^{\top}$ and the averaging operator ${\mathbf{W}^t = \mathbf{I}_n - \frac{1}{2}(e_i - e_j)(e_i - e_j)^{\top}}$, where $i$ and $j$ are the nodes awaken at iteration $t$. An iteration of Algorithm~\ref{alg:deleda} can be reformulated as follows:
\begin{equation}
  \mathbf{S}^{t + 1} = (1 - \rho_{t + 1})\mathbf{W}^{t + 1}\mathbf{S}^t + \rho_{t + 1} \mathbf{G}^{t + 1}, \nonumber
\end{equation}
where $\mathbf{G}^{t + 1}$ is the matrix of expectations in \eqref{eqn:G-OEM}. Recursively, one can obtain:
\begin{equation}
  \mathbf{S}^t = \sum_{r = 0}^t (1 - \rho_t)\ldots(1 - \rho_{r+1})\rho_r \mathbf{W}^t \ldots \mathbf{W}^r \mathbf{G}^r. \nonumber
\end{equation}
Using properties of averaging matrices\footnote{See \cite{boyd2006a, colin2016gossip} for more detailed analysis of averaging matrices.}, one can rearrange the above expression and write:
\begin{align}
  \bbE \| \mathbf{S}^t - \overline{s}^t \1_n^{\top} \| 
  &\leq \sum_{r = 0}^t (1 - \rho_t)\ldots(1 - \rho_{r+1}) \rho_r \bbE \left\|(\mathbf{W}^t - \1_{n \times n} ) \ldots (\mathbf{W}^r - \1_{n \times n}) \mathbf{G}^r \right\| \nonumber\\
  &\leq \sum_{r = 0}^t \rho_r \lambda_2^{\frac{t - r}{2}} \bbE \|\mathbf{G}^r\|, \label{ineq:spectral-gap}
\end{align}
where $\lambda_2$ is the second largest eigenvalue of $\bbE[\mathbf{W^t}]$. If the graph $\mathcal{G}$ is connected and non-bipartite, one can show that $0 < \lambda_2 < 1$. Therefore, when $(\rho_t)_{t \geq 0}$ tends towards 0 and provided that $\bbE\|\mathbf{G}_r\|$ is bounded, the expected gap between nodes iterates and the average $\overline{s}^t$ goes to 0. In practice, this last assumption holds for instance when the Dirichlet parameter $\alpha$ is either fixed or lower-bounded, \emph{i.e.,} $\alpha > r > 0$. Also, as evidenced by inequality~\eqref{ineq:spectral-gap}, one should expect {\sc DeLeDA} to converge faster on a graph with high spectral gap, as it is proportional to $1 - \lambda_2$~\cite{colin2016gossip, boyd2006a}.
\begin{remark}
  Note that this reasoning holds in the synchronous setting. In the fully asynchronous setting, one needs to take into account the nodes degrees and adapt the update rule~\eqref{eqn:G-OEM} for the algorithm to converge to the true objective, as detailed for instance in~\cite{colin2016gossip}.
\end{remark}

\section{Numerical experiments}
\label{sec:numer-exper}
\paragraph{Dataset.}
We perform experiments on a synthetic graph of $n$=50 nodes. Given these nodes, we consider the complete graph ($|\mathcal{E}|=1225$ edges) and a graph where the edges are generated through a Watts-Strogatz model~\cite{watts1998collective}, with $100$ edges and rewiring probability $p = 0.3$. We attach 20 documents at each node, generated through LDA generative process, with a vocabulary size $V=100$ and $K=5$ topics. The length of each document is drawn from a Poisson with parameter 10. We compare our algorithm to --- centralized --- G-OEM applied to the 1000 synthetic documents of the graph with a batch size of 20 documents. In particular, at each iteration of the G-OEM algorithm we choose uniformly at random 20 documents among the 1000 of the total dataset, and update $\eta$ according to the G-OEM process. In practice, we update $\beta$ at each iteration and let $\alpha=\alpha^*$ fixed, as often done in previous work \cite{OnlineLDA,mimno2012sparse,Patterson2013,StreamingGibbs2016}.

\paragraph{Evaluation.}
We evaluate our method by computing the likelihood on held-out documents, that is $p(X|\eta)$ for any test document $X$. For LDA, the likelihood is intractable to compute. We approximate $p(X|\eta)$ with the ``left-to-right'' evaluation algorithm \citep{wallachEvaluation} applied to each test document.
In the following, we present results in terms of log-perplexity, defined as the opposite of the log-likelihood~${LP=-\log p(X|\eta)}$. The lower the log-perplexity, the better the corresponding model. In our experiments, we  compute the average test log-perplexity over test documents. More precisely, we display the relative error $(LP / LP^* - 1)$, where ${LP^*}$ is the average log-perplexity of test documents computed with optimal parameters $\eta^*$ that generated the data.

We also compare the performance in terms of distance to the optimal topic matrix $\beta^*$, which generated the documents. As the performance is invariant to topic permutation in LDA, we compute the following distance between the infered parameter $\beta$ and optimal parameter $\beta^*$:
\[
D(\beta,\beta^*) = \min_{M\in\mathbb{R}^{K\times K}} \frac{\Vert M \beta - \beta^*\Vert_F}{\Vert\beta^*\Vert_F} = \frac{\Vert \beta^*\beta^\top(\beta\beta^\top)^{-1} - \beta^*\Vert_F}{\Vert\beta^*\Vert_F},
\]
which is invariant by topic permutation (\ie by permutation of rows of $\beta$).

\begin{figure}[t]
  \centering
  \begin{subfigure}[t]{.42\textwidth}
    \centering
    \includegraphics[width=\textwidth]{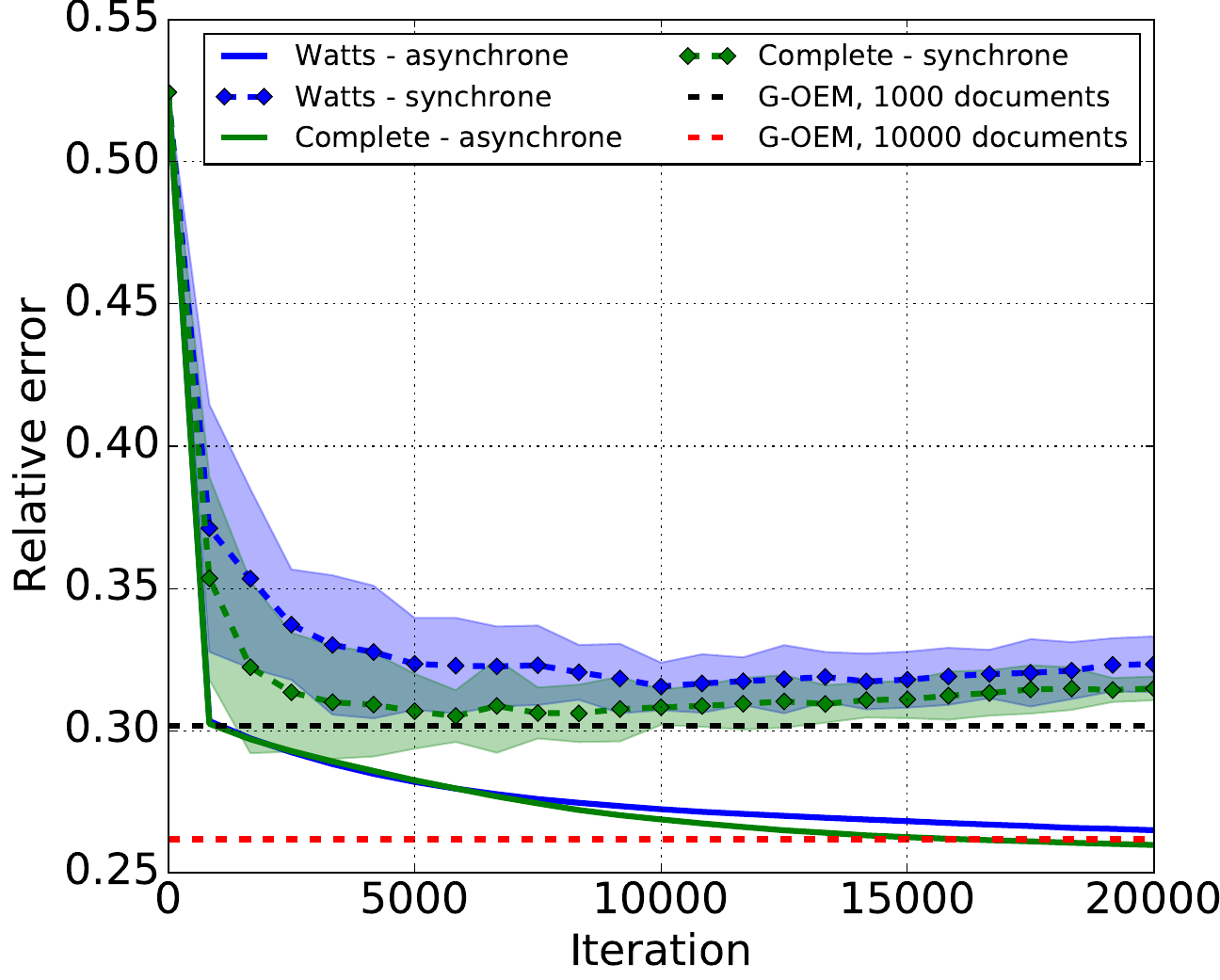} 
    \caption{Relative error $\left(LP-LP^*\right)$.}
    \label{fig:perpl}
  \end{subfigure}
  \hspace{1cm}
  \begin{subfigure}[t]{.42\textwidth}
    \centering
    \includegraphics[width=\textwidth]{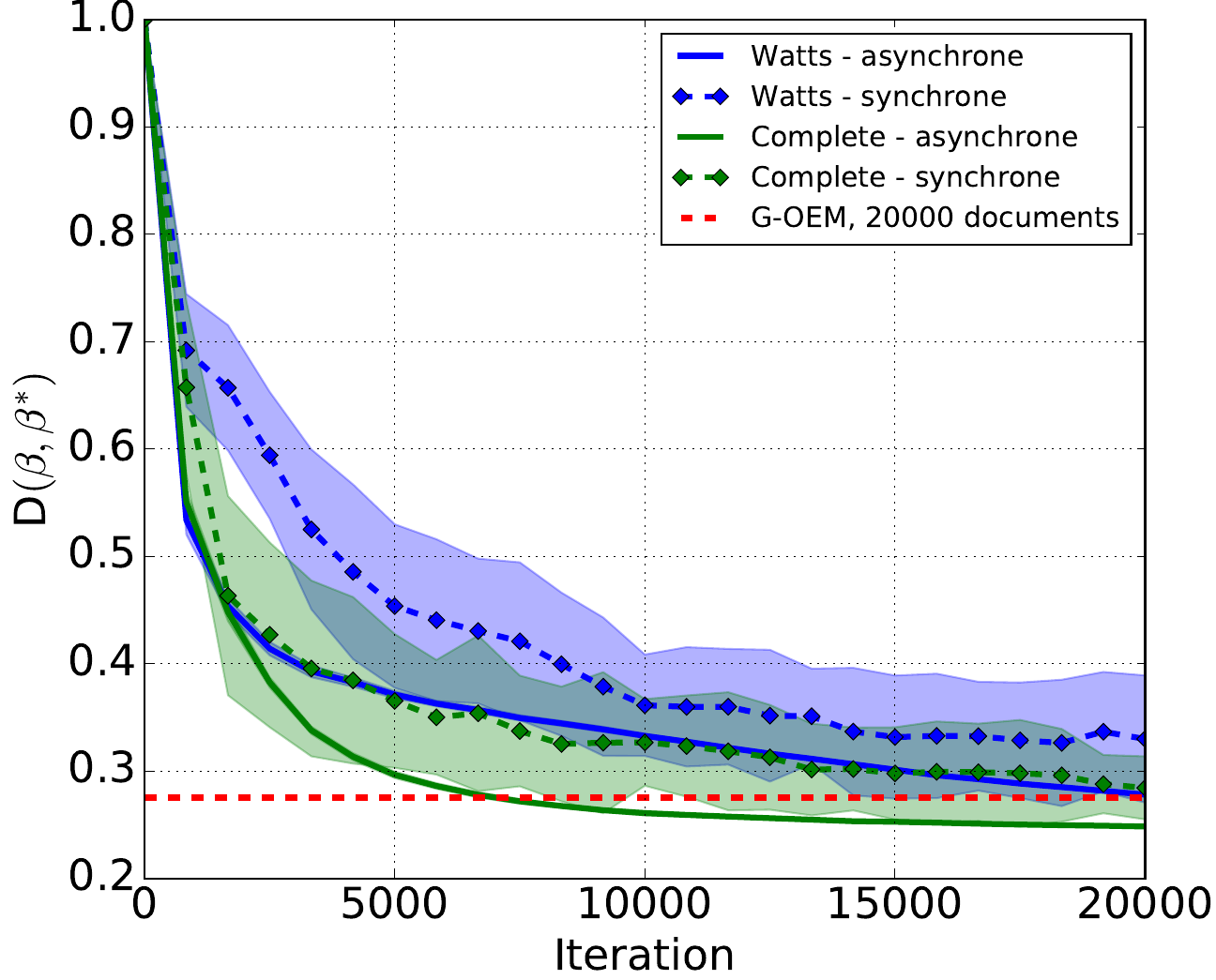} 
    \caption{Average distance to the optimal parameter $\beta^*$.}
    \label{fig:dist}
  \end{subfigure}
  \caption{Distance of log-perplexity (\subref{fig:perpl}) and parameters (\subref{fig:dist}) as a function of the iterations.}
  \label{fig:results}
\end{figure}

Results for asynchrone and synchrone method are presented in Figure~\ref{fig:results}. We observe that our method converges to the same state than G-OEM after enough iterations. In particular, each agent is able to recover the topic matrix that generated all the documents without any direct access to documents of the other nodes.

As expected, {\sc DeLeDA} converges faster on a perfectly connected network, even though the ``well-connected'' design of the Watts-Strogatz network allows it to stay close to the complete graph, perplexity-wise. Finally, the synchronous version of {\sc DeLeDA} appears to converge slower than the asynchronous one. This is an overfitting issue: synchronous computations lead each node to perform roughly $n/2$ local updates between two averaging steps, whereas asynchronous computations force nodes to update only once between two averaging steps.

\newpage
\bibliography{bib}

\begin{thebibliography}{10}

\bibitem{nedic2009distributed}
Angelia Nedic and Asuman Ozdaglar.
\newblock Distributed subgradient methods for multi-agent optimization.
\newblock {\em IEEE Transactions on Automatic Control}, 54(1):48--61, 2009.

\bibitem{nedic2011asynchronous}
Angelia Nedi{\'c}.
\newblock Asynchronous broadcast-based convex optimization over a network.
\newblock {\em Automatic Control, IEEE Transactions on}, 56(6):1337--1351,
  2011.

\bibitem{agarwal2010distributed}
Alekh Agarwal, Martin~J Wainwright, and John~C Duchi.
\newblock Distributed dual averaging in networks.
\newblock In {\em Advances in Neural Information Processing Systems}, pages
  550--558, 2010.

\bibitem{colin2016gossip}
I.~Colin, A.~Bellet, J.~Salmon, and S.~Cl{\'e}men{\c{c}}on.
\newblock Gossip dual averaging for decentralized optimization of pairwise
  functions.
\newblock {\em Proc. ICML}, 2016.

\bibitem{LDA}
D.~M. Blei, A.~Y. Ng, and M.~I. Jordan.
\newblock Latent {D}irichlet allocation.
\newblock {\em JMLR}, 3:993--1022, 2003.

\bibitem{MeOnlineEM2016}
C.~Dupuy and F.~Bach.
\newblock Online but accurate inference for latent variable models with local
  gibbs sampling.
\newblock {\em arXiv preprint arXiv:1603.02644}, 2016.

\bibitem{OnlineEM}
O.~Capp{\'e} and E.~Moulines.
\newblock Online {EM} algorithm for latent data models.
\newblock {\em Journal of the Royal Statistical Society}, 71(3):593--613, 2009.

\bibitem{OnlineLDA}
M.~D. Hoffman, D.~M. Blei, and F.~R. Bach.
\newblock Online learning for latent {D}irichlet allocation.
\newblock {\em Adv. NIPS}, 2010.

\bibitem{GibbsLDA}
T.~L. Griffiths and M.~Steyvers.
\newblock Finding scientific topics.
\newblock {\em Proceedings of the National Academy of Sciences}, 101(suppl
  1):5228--5235, 2004.

\bibitem{SinglePassLDA}
I.~Sato, K.~Kurihara, and H.~Nakagawa.
\newblock Deterministic single-pass algorithm for {LDA}.
\newblock {\em Adv. NIPS}, 2010.

\bibitem{mimno2012sparse}
D.~Mimno, M.~Hoffman, and D.~Blei.
\newblock Sparse stochastic inference for latent {D}irichlet allocation.
\newblock {\em Proc. ICML}, 2012.

\bibitem{ExpFam1998}
E.~L. Lehmann and G.~Casella.
\newblock {\em Theory of point estimation}, volume~31.
\newblock Springer Science \& Business Media, 1998.

\bibitem{EM1977}
A.~P. Dempster, N.~M. Laird, and D.~B. Rubin.
\newblock Maximum likelihood from incomplete data via the {EM} algorithm.
\newblock {\em Journal of the royal statistical society. Series B
  (methodological)}, 39(1):1--38, 1977.

\bibitem{boyd2006a}
Stephen Boyd, Arpita Ghosh, Balaji Prabhakar, and Devavrat Shah.
\newblock {R}andomized gossip algorithms.
\newblock {\em IEEE Trans. Inf. Theory}, 52(6):2508--2530, 2006.

\bibitem{watts1998collective}
Duncan~J Watts and Steven~H Strogatz.
\newblock Collective dynamics of ‘small-world’networks.
\newblock {\em Nature}, 393(6684):440--442, 1998.

\bibitem{Patterson2013}
S.~Patterson and Y.~W. Teh.
\newblock Stochastic gradient {R}iemannian langevin dynamics on the probability
  simplex.
\newblock {\em Adv. NIPS}, 2013.

\bibitem{StreamingGibbs2016}
Y.~Gao, J.~Chen, and J.~Zhu.
\newblock Streaming gibbs sampling for {LDA} model.
\newblock {\em arXiv preprint arXiv:1601.01142}, 2016.

\bibitem{wallachEvaluation}
H.~M. Wallach, I.~Murray, R.~Salakhutdinov, and D.~Mimno.
\newblock Evaluation methods for topic models.
\newblock {\em Proc. ICML}, 2009.

\end{thebibliography}

\end{document}